\documentclass{article}
\usepackage[preprint]{neurips_2026}
\usepackage[utf8]{inputenc}
\usepackage[T1]{fontenc}
\usepackage{hyperref}
\usepackage{url}
\usepackage{booktabs}
\usepackage{amsfonts,amssymb}
\usepackage{amsmath,amsthm,mathtools}
\usepackage{nicefrac}
\usepackage{microtype}
\usepackage{xcolor}
\usepackage{algorithm}
\usepackage{algpseudocode}
\usepackage{tabularx}
\usepackage{pifont}
\usepackage{placeins}
\definecolor{citationgreen}{rgb}{0.64,0.99,0.65}
\definecolor{checkgreen}{RGB}{35,139,69}
\definecolor{crossred}{RGB}{202,52,51}

\hypersetup{
  hypertexnames=false,
  colorlinks=false,
  linkcolor=black,
  citecolor=citationgreen,
  citebordercolor=citationgreen,
  urlcolor=black
}

\title{Unifying Embodied Foundation Models via a Model-to-Policy Formulation}
\author{
  Xinyi Hong, Mingyang Ren\thanks{Corresponding author} \\
  Department of Statistics, School of Mathematical Sciences \\
  Shanghai Jiao Tong University \\
  \texttt{xinyi.hong@sjtu.edu.cn}, \texttt{mingyangren@sjtu.edu.cn}
}
\begin{document}
\maketitle
\begin{abstract}
 Vision-language-action (VLA) models and world-action models (WAMs) have emerged as promising approaches to robot learning and control. These models integrate multimodal perception and action generation, but differ in their architectures, output representations, and learning objectives. In this paper, we present a unified, architecture-agnostic probabilistic formulation for embodied foundation models based on stochastic kernels. Our formulation has two key components: (i) we characterize embodied foundation models through conditional output distributions and action-realization kernels, which together induce task-conditioned robot policies; and (ii) we connect pretraining and post-training to policy deployment, accommodating model-specific pretraining objectives and offline or online post-training strategies. We instantiate the formulation for OpenVLA and DreamZero, covering autoregressive action-token prediction and joint video-action generation.
\end{abstract}

\section{Introduction}
Robot learning is moving from task-specific controllers toward embodied AI systems that learn reusable perception-to-action policies across tasks and embodiments~\citep{openxembodiment2024rtx,octomodelteam2024octoopensourcegeneralistrobot}. At its current stage, embodied AI largely adapts the broader foundation-model development pipeline to physical interaction~\citep{driess2023palme}. This pipeline combines a shared architectural core with three stages: pre-training, post-training, and deployment. The shared architectural core is typically based on the Transformer family, with modality-specific encoders, tokenizers, or output heads added (if needed)~\citep{vaswani2017attention,brown2020language,geminiteam2024gemini15,xu2025qwen25omni,qwenteam2026qwen35omni}. First, pre-training learns general representations from large-scale corpora and multimodal collections through model-specific objectives~\citep{brown2020language,geminiteam2024gemini15,qwenteam2026qwen35omni}. Specifically for LLMs, the resulting next-token loss follows predictable power-law relations in model size, data, and compute (similarly rigorous scaling laws are less established for other foundation models)~\citep{kaplan2020scaling,hoffmann2022training,yang2025qwen3}. Second, post-training adapts the pretrained model to follow instructions, align with human or verifiable preferences, and improve downstream task performance~\citep{wei2021finetuned,ouyang2022training}. Common approaches include supervised fine-tuning (SFT) and preference optimization (e.g., DPO and KTO)~\citep{ouyang2022training,rafailov2023direct,ethayarajh2024kto}. Other routes use reinforcement learning (e.g., RLHF and RLVR, often implemented with PPO or GRPO) or on-policy distillation (OPD)~\citep{ouyang2022training,schulman2017proximal,shao2024deepseekmath,deepseekai2025deepseekr1,agarwal2024onpolicy}. Finally, deployment turns the trained model into a reusable inference and reasoning core. The above foundation-model pipeline has shown broad effectiveness across scales and modalities, underpinning recent language and reasoning models (e.g., GPT-5.6 and DeepSeek-V4), multimodal or omnimodal models such as MiniMax M3 and Qwen3.5-Omni, and generative models including FLUX.2~\citep{openai2026gpt56,deepseekai2026deepseekv4,qwenteam2026qwen35omni}. The resulting models can also serve as reasoning and perception cores for agents~\citep{hong2026tools}. Such agents support tool use, coding, digital-environment interaction, and multi-agent coordination~\citep{yao2023react,schick2023toolformer,yang2024sweagent,wu2024autogen}. Can the success of large foundation models be carried into the physical world? The above non-embodied foundation models generally learn a mapping \(f_\theta:\mathcal X\to\mathcal Y\) from input data to predictions or generated content~\citep{hong2026hyvint}. However, for robot learning, embodied foundation models aim to induce a task-conditioned policy $\pi_\theta:\mathcal T\times\mathcal C\to\mathcal P(\mathcal A)$ that maps the available interaction history and task condition to a probability distribution over robot actions. Fitting a fixed data distribution is insufficient, since a policy must also perform well on the trajectories induced by its own actions~\citep{kaelbling1998planning,ross2011reduction,brohan2023rt2visionlanguageactionmodelstransfer}.  
Vision-language-action (VLA) models pursue this policy-learning objective by combining visual observations and language instructions to generate robot actions. The Robotics Transformer (RT) series develops this approach from RT-1's discretized action prediction to RT-2's transfer of internet-scale visual-semantic knowledge through autoregressive action-token prediction~\citep{brohan2023rt1,brohan2023rt2visionlanguageactionmodelstransfer}. OpenVLA likewise adapts a pretrained vision-language backbone with an action-token cross-entropy objective, while $\pi_0$ couples such a backbone to a flow-matching action expert that generates continuous action chunks~\citep{kim2025openvla,black2024pi}. Scaling their robot pre-training is constrained by the difficulty of collecting real robot trajectories across diverse tasks, environments, and embodiments~\citep{openxembodiment2024rtx,black2024pi}. Video pre-training offers a complementary route to learning motion and scene dynamics from internet-scale data without robot action labels, reducing reliance on embodiment-specific trajectories for acquiring these priors~\citep{assran2025vjepa2selfsupervisedvideo,ye2026world}. For control, the value of this pre-training lies in acquiring transferable regularities of physical interaction that support action, rather than visual realism alone~\citep{assran2025vjepa2selfsupervisedvideo}. Such priors support control through predictive world models, as in UniPi, which uses inverse dynamics to extract robot actions from generated video plans~\citep{ha2018world,du2023learning}. World-action models (WAMs) couple these visual predictions with action generation under language conditioning, as in DreamZero, which adapts a pretrained video diffusion backbone to jointly generate future video and robot actions~\citep{ye2026world}. Conditioned on visual history and proprioception, DreamZero uses flow matching within chunks and autoregressive generation across chunks~\citep{ye2026world}. Real robot trajectories ground these pretrained representations in embodiment-specific commands and physical transitions, supporting transfer across tasks, environments, and embodiments~\citep{openxembodiment2024rtx,physicalintelligence2025pi05}. Adapting these representations to downstream control can also change the model's outputs, as $\pi_{0.5}$ adds a continuous action expert during post-training after discrete-token pre-training~\citep{physicalintelligence2025pi05}. The resulting outputs must then be converted into executable actions through decoding and action selection or scheduling~\citep{kim2025openvla,black2024pi,ye2026world}. These actions change the observations conditioning subsequent predictions, so the deployed policy depends on both the learned output distribution and its realization in closed-loop interaction. A common description must therefore connect these two components with the objectives and data used in pre-training and post-training.

Motivated by this connection, we develop an architecture-agnostic probabilistic formulation based on stochastic kernels. To the best of our knowledge, this is the first formulation that explicitly connects model output distributions and action realization with pre-training, post-training, and closed-loop robot interaction for both VLAs and WAMs. We instantiate it for OpenVLA and DreamZero.

\section{Task-Conditioned Robot Interaction Setting}
We consider a robot interaction setting in which a robot observes its environment and takes actions to complete a given task. Let \(T\) denote the number of action steps, indexed by \(t\in\{0,\ldots,T-1\}\), with \(o_t\in\mathcal O\) denoting the observation available at step \(t\) and \(a_t\in\mathcal A\) the action selected at that step, where \(\mathcal O\) is the observation space and \(\mathcal A\) is the action space. Specifically, given a task condition \(c\in\mathcal C\) and an initial observation \(o_0\), the robot selects \(a_t\) based on the available trajectory \(\tau_{\le t}=(o_0,a_0,\ldots,o_t)\in\mathcal T\), where \(\mathcal C\) is the space of task conditions and \(\mathcal T\) is the space of partial trajectories. The robot then receives \(o_{t+1}\) and repeats this process with the new observation for \(t=0,\ldots,T-1\). After \(T\) action steps, the resulting complete trajectory is \(\tau=(o_0,a_0,\ldots,a_{T-1},o_T)\). Within this setting, we aim to model task-conditioned action selection through a policy
$
\pi_\theta:\mathcal T\times\mathcal C\to\mathcal P(\mathcal A),
$
where \(\mathcal P(\mathcal A)\) denotes the space of probability distributions over actions and \(\theta\) denotes the learnable parameters. At each step \(t\), given \(\tau_{\le t}\) and \(c\), the robot samples its next action through \(a_t\sim\pi_\theta(\cdot\mid\tau_{\le t},c)\).
\section{Unified Formulation for Embodied Foundation Models}
We present a unified, architecture-agnostic formulation of how an embodied foundation model induces $\pi_\theta(\cdot\mid\tau_{\le t},c)$ through pre-training and post-training. Firstly, we denote the embodied foundation model by $f_\theta:\mathcal T\times\mathcal C\to\mathcal P(\mathcal Y)$, where $\mathcal Y$ is the output space and $\mathcal P(\mathcal Y)$ denotes the space of probability measures on $\mathcal Y$. Here, \(f_\theta\) incorporates the specified sampling, decoding, planning, and inference procedures, including asynchronous generation when applicable. At each step $t$, the output $y_t\in\mathcal Y$ has conditional distribution $y_t\mid(\tau_{\le t},c)\sim f_\theta(\cdot\mid\tau_{\le t},c)$. That is, $f_\theta(E\mid\tau_{\le t},c)=\Pr_\theta(y_t\in E\mid\tau_{\le t},c)$ for every measurable set $E\subseteq\mathcal Y$. To obtain $a_t$ from $y_t$, we define a measurable action-realization kernel $q_\theta:\mathcal Y\times\mathcal T\times\mathcal C\to\mathcal P(\mathcal A)$ and sample $a_t\sim q_\theta(\cdot\mid y_t,\tau_{\le t},c)$. The action-realization kernel $q_\theta$ maps model outputs to executable actions, incorporating action selection and scheduling procedures when applicable. By specifying the output space $\mathcal Y$ and the corresponding kernel $q_\theta$, the unified formulation accommodates several representative embodied model architectures as follows. The expressions below hold for every measurable set $B\subseteq\mathcal A$, where $\mathbf 1\{\cdot\}$ denotes the indicator function:
\begin{itemize}
    \item[(i)] Direct action outputs: $\mathcal Y=\mathcal A$, $y=a$, and $q_\theta(B\mid y,\tau_{\le t},c)=\mathbf 1\{a\in B\}$ (e.g., RoboFlamingo~\citep{li2024vision}).
\item[(ii)] Joint video and action chunks:
$\mathcal Y=\mathcal I^F\times\mathcal A^H$ and
$y=(\mathbf I',\mathbf a')$, where $\mathcal I$ is the image space and $F,H\ge1$ are the numbers of predicted frames and actions, respectively  (e.g., DreamZero~\citep{ye2026world}). At step $t$, $y$ denotes the available joint chunk, which may have been generated at an earlier step. The action-realization kernel $q_\theta(B\mid y,\tau_{\le t},c)$ specifies the conditional probability that the executable action selected from the available chunk belongs to $B$, incorporating action scheduling and smoothing when applicable.
    \item[(iii)] Action tokens with decoded actions: $\mathcal Y=\mathcal V^\ast\times\mathcal A$, $y=(s,a)$, and $q_\theta(B\mid y,\tau_{\le t},c)=\mathbf 1\{a\in B\}$, where $\mathcal V^\ast$ is the set of finite token sequences over a vocabulary $\mathcal V$, and $a$ is obtained by decoding $s$ through the procedure included in $f_\theta$ (e.g., OpenVLA~\citep{kim2025openvla}).
    \item[(iv)] Action chunks~\citep{zhao2023learning,chi2025diffusion}: $\mathcal Y=\mathcal A^H$, $y=\mathbf a$, and $q_\theta(B\mid y,\tau_{\le t},c)=\mathbf 1\{a_0\in B\}$, where $H\ge1$ is a prediction horizon and $\mathbf a=(a_0,\ldots,a_{H-1})$ is an action chunk with indices local to the chunk. This realization executes the first predicted action before replanning (e.g., GR00T N1~\citep{bjorck2025gr00t}).
    \item[(v)] Latent actions with decoded actions: $\mathcal Y=\mathcal Z\times\mathcal A$, $y=(z,a)$, and $q_\theta(B\mid y,\tau_{\le t},c)=\mathbf 1\{a\in B\}$, where $\mathcal Z$ is a latent action space and $a$ is obtained by decoding $z$ through the procedure included in $f_\theta$ (e.g., VQ-VLA~\citep{wang2025vq}).
    \item[(vi)] Video plans with extracted actions: $\mathcal Y=\mathcal O^H\times\mathcal A$, $y=(\mathbf o',a)$, and $q_\theta(B\mid y,\tau_{\le t},c)=\mathbf 1\{a\in B\}$, where $H\ge1$ is a prediction horizon, $\mathbf o'=(o'_1,\ldots,o'_H)$ is a video plan with $o'_j\in\mathcal O$, and $a$ is obtained from this plan through the action-extraction procedure included in $f_\theta$~\citep{du2023learning} (e.g., LVP~\citep{chen2025large}).
\end{itemize}
\paragraph{Pre-training.}
We consider offline pre-training as follows. Given a pre-training dataset $\mathcal D_{\mathrm{pre}}=\{z_i^{\mathrm{pre}}\}_{i=1}^{N_{\mathrm{pre}}}$, we write the corresponding empirical pre-training loss as
\begin{equation}
\widehat{\mathcal L}_{\mathrm{pre}}(\theta)
=
\frac{1}{N_{\mathrm{pre}}}
\sum_{i=1}^{N_{\mathrm{pre}}}
\ell_{\mathrm{pre}}(\theta,z_i^{\mathrm{pre}}),
\label{eq:pretr}
\end{equation}
where $\ell_{\mathrm{pre}}(\theta,z)$ denotes the per-example loss specified by the chosen pre-training method, and $z$ denotes a training example. We estimate $\theta$ by optimizing $\widehat{\mathcal L}_{\mathrm{pre}}(\theta)$ and set $\theta_{\mathrm{pre}}\leftarrow\theta$.
\paragraph{Post-training.}
To adapt $f_{\theta_{\mathrm{pre}}}$ and $q_{\theta_{\mathrm{pre}}}$ for robot action selection, let $\mathcal D_{\mathrm{off}}=\{z_i^{\mathrm{off}}\}_{i=1}^{N_{\mathrm{off}}}$ denote a fixed offline dataset and $\mathcal D_{\mathrm{on}}^{(k)}=\{z_i^{\mathrm{on},(k)}\}_{i=1}^{N_{\mathrm{on}}^{(k)}}$ denote online experience available at iteration $k$. Online experience is collected through interaction with real or simulated environments using current or previously used policies and may be accumulated or refreshed across iterations~\citep{ross2011reduction,ball2023efficient}. We have the following empirical loss:
\begin{equation}
\widehat{\mathcal L}_{\mathrm{post}}^{(k)}(\theta)
=
\frac{1-\alpha_k}{N_{\mathrm{off}}}
\sum_{i=1}^{N_{\mathrm{off}}}
\ell_{\mathrm{off}}^{(k)}(\theta,z_i^{\mathrm{off}})
+
\frac{\alpha_k}{N_{\mathrm{on}}^{(k)}}
\sum_{i=1}^{N_{\mathrm{on}}^{(k)}}
\ell_{\mathrm{on}}^{(k)}(\theta,z_i^{\mathrm{on},(k)}),
\label{eq:postr}
\end{equation}
where $\ell_{\mathrm{off}}^{(k)}$ and $\ell_{\mathrm{on}}^{(k)}$ are the corresponding per-example losses specified by the chosen post-training method and $\alpha_k\in[0,1]$ controls their relative weighting. Specifically, $\alpha_k=0$ uses only offline data, $\alpha_k=1$ uses only online data, and $\alpha_k\in(0,1)$ combines both. These losses update the learnable parameters of \(f_\theta\) and, where applicable, \(q_\theta\), which together determine the induced policy \(\pi_\theta\). Starting from $\theta^{(0)}=\theta_{\mathrm{pre}}$, we optimize $\widehat{\mathcal L}_{\mathrm{post}}^{(k)}$ iteratively and obtain $\theta_{\mathrm{post}}=\theta^{(K)}$ after $K$ iterations.
\paragraph{Deployment.}
Combining outputs from $f_{\theta_{\mathrm{post}}}$ with the action-realization kernel $q_{\theta_{\mathrm{post}}}$ gives the deployed policy. Depending on the specific implementation, action realization may be deterministic or stochastic, as described below for any measurable action set $B\subseteq\mathcal A$:
\begin{itemize}
    \item[(i)] Deterministic action realization: $q_{\theta_{\mathrm{post}}}$ concentrates on a single action given $y$, $\tau_{\le t}$, and $c$, so $q_{\theta_{\mathrm{post}}}(B\mid y,\tau_{\le t},c)\in\{0,1\}$. Conditional on $\tau_{\le t}$ and $c$, the events $a_t\in B$ and $q_{\theta_{\mathrm{post}}}(B\mid y_t,\tau_{\le t},c)=1$ agree almost surely. Consequently,
    \begin{equation}
    \begin{aligned}
    \pi_{\theta_{\mathrm{post}}}(B\mid\tau_{\le t},c)
    &:=\Pr_{\theta_{\mathrm{post}}}(a_t\in B\mid\tau_{\le t},c)\\
    &=\Pr_{\theta_{\mathrm{post}}}
    \bigl(q_{\theta_{\mathrm{post}}}(B\mid y_t,\tau_{\le t},c)=1
    \mid\tau_{\le t},c\bigr)\\
    &=\int_{\mathcal Y}
    \mathbf 1\{q_{\theta_{\mathrm{post}}}(B\mid y,\tau_{\le t},c)=1\}
    f_{\theta_{\mathrm{post}}}(\mathrm d y\mid\tau_{\le t},c).
    \end{aligned}
    \label{eq:deter}
    \end{equation}
    \item[(ii)] Stochastic action realization: $q_{\theta_{\mathrm{post}}}$ need not concentrate on a single action given $y$, $\tau_{\le t}$, and $c$. Conditioning on $y_t$ and applying the law of total expectation gives
    \begin{equation}
    \begin{aligned}
    \pi_{\theta_{\mathrm{post}}}(B\mid\tau_{\le t},c)
    &:=\Pr_{\theta_{\mathrm{post}}}(a_t\in B\mid\tau_{\le t},c)\\
    &=\mathbb E_{\theta_{\mathrm{post}}}
    \bigl[q_{\theta_{\mathrm{post}}}(B\mid y_t,\tau_{\le t},c)
    \mid\tau_{\le t},c\bigr]\\
    &=\int_{\mathcal Y}
    q_{\theta_{\mathrm{post}}}(B\mid y,\tau_{\le t},c)
    f_{\theta_{\mathrm{post}}}(\mathrm d y\mid\tau_{\le t},c).
    \end{aligned}
    \label{eq:polic}
    \end{equation}
    Although Eq.~\ref{eq:polic} generalizes Eq.~\ref{eq:deter} mathematically, we explicitly separate them to highlight the practical distinction between deterministic decoding and stochastic sampling.
\end{itemize}
In either case, sampling $y_t\sim f_{\theta_{\mathrm{post}}}(\cdot\mid\tau_{\le t},c)$, then sampling $a_t\sim q_{\theta_{\mathrm{post}}}(\cdot\mid y_t,\tau_{\le t},c)$ and executing it implements $\pi_{\theta_{\mathrm{post}}}$.
\begin{algorithm}[t]
\caption{Embodied Foundation Models: End-to-End Learning and Deployment}
\begin{algorithmic}[1]
\Statex \textbf{Input:} Model $f_\theta$, action-realization kernel
$q_\theta$, initial parameters $\theta$,
pre-training data $\mathcal D_{\mathrm{pre}}$,
offline data $\mathcal D_{\mathrm{off}}$,
initial online experience $\mathcal D_{\mathrm{on}}^{(0)}$,
losses $\ell_{\mathrm{pre}}$,
$\{\ell_{\mathrm{off}}^{(k)},\ell_{\mathrm{on}}^{(k)}\}_{k=0}^{K-1}$,
mixing weights $\{\alpha_k\}_{k=0}^{K-1}$,
post-training iterations $K$,
environment access, task condition $c$,
initial observation $o_0$, deployment horizon $T$
\Statex \textbf{Output:} Parameters $\theta_{\mathrm{post}}$
and trajectory $\tau$

\Statex \textbf{\textit{Pre-training phase}}

\State Optimize $\theta$ by minimizing
Eq.~\eqref{eq:pretr}
\State Set $\theta_{\mathrm{pre}} \gets \theta$

\Statex \textbf{\textit{Post-training phase}}
\State Initialize $\theta^{(0)} \gets \theta_{\mathrm{pre}}$
\For{$k=0,\ldots,K-1$}
    \If{$\alpha_k>0$}
        \State Collect or refresh $\mathcal D_{\mathrm{on}}^{(k)}$
        through interaction under $\pi_{\theta^{(k)}}$
    \EndIf
    \State Obtain $\theta^{(k+1)}$ by optimizing
    Eq.~\eqref{eq:postr} from $\theta^{(k)}$
\EndFor
\State Set $\theta_{\mathrm{post}} \gets \theta^{(K)}$

\Statex \textbf{\textit{Deployment phase}}
\State Initialize $\tau_{\le 0} \gets (o_0)$
\For{$t=0,\ldots,T-1$}
    \State Sample $y_t \sim
    f_{\theta_{\mathrm{post}}}(\cdot\mid\tau_{\le t},c)$
    \State Sample $a_t \sim
   q_{\theta_{\mathrm{post}}}(\cdot\mid y_t,\tau_{\le t},c)$
    \State Execute $a_t$ and receive $o_{t+1}$
    \State Update $\tau_{\le t+1} \gets
    (\tau_{\le t},a_t,o_{t+1})$
\EndFor
\State \Return $\theta_{\mathrm{post}}$ and
$\tau \gets \tau_{\le T}$
\end{algorithmic}
\end{algorithm}

\section{Vision-Language-Action (VLA) Model}
\label{sec:vlmod}

\paragraph{Model Architecture.} We specialize the unified formulation to the OpenVLA architecture~\citep{kim2025openvla}. Specifically, we represent the model as $f_\theta:\mathcal O\times\mathcal C\to\mathcal P(\mathcal Y)$, where $\mathcal O$ is the image observation space, $\mathcal C$ is the language instruction space, $\mathcal Y=\mathcal V^\ast\times\mathcal A$ is the output space, $\mathcal V^\ast$ denotes the set of finite token sequences over the vocabulary $\mathcal V$, $\mathcal A\subseteq\mathbb R^7$ is the robot action space, and $\mathcal P(\mathcal Y)$ denotes the space of probability distributions over $\mathcal Y$. Unlike the general formulation, OpenVLA conditions only on the current image $o_t$ and language instruction $c$, so $f_\theta(\cdot\mid\tau_{\le t},c)=f_\theta(\cdot\mid o_t,c)$. Let $\theta=(\theta_{\mathrm D},\theta_{\mathrm S},\theta_{\mathrm P},\theta_{\mathrm L})$ denote the parameters of DINOv2~\citep{oquab2023dinov2}, SigLIP~\citep{zhai2023sigmoid}, the MLP projector, and the language model, respectively. 

First, at step $t$, given an input image $o_t\in\mathcal O$, the visual encoders $\mathrm{DINOv2}_{\theta_{\mathrm D}}:\mathcal O\to\mathbb R^{256\times1024}$ and $\mathrm{SigLIP}_{\theta_{\mathrm S}}:\mathcal O\to\mathbb R^{256\times1152}$ extract features for 256 spatially aligned patches, with their respective $224\times224$ image preprocessing included in these encoders. Then, we horizontally concatenate their outputs as $[\mathrm{DINOv2}_{\theta_{\mathrm D}}(o_t),\mathrm{SigLIP}_{\theta_{\mathrm S}}(o_t)]\in\mathbb R^{256\times2176}$. The concatenated outputs are then fed into an MLP projector $\mathrm{MLP}_{\theta_{\mathrm P}}:\mathbb R^{256\times2176}\to\mathbb R^{256\times4096}$, yielding the visual token embeddings $H_t^{\mathrm{img}}=\mathrm{MLP}_{\theta_{\mathrm P}}([\mathrm{DINOv2}_{\theta_{\mathrm D}}(o_t),\mathrm{SigLIP}_{\theta_{\mathrm S}}(o_t)])\in\mathbb R^{256\times4096}$. 

Second, given a language instruction $c\in\mathcal C$, the tokenizer $\mathrm{Tok}:\mathcal C\to\mathcal V^\ast$ applies the fixed instruction-prompt format and tokenizer, producing $\mathrm{Tok}(c)=(t_1,\ldots,t_m)\in\mathcal V^m$, where $m$ is the prompt length and $t_i$ is the $i$-th token, with $t_1$ being the beginning-of-sequence token. These tokens are then mapped into the language-model embedding space, yielding the language token embeddings $H_t^{\mathrm{lang}}=\mathrm{Embed}_{\theta_{\mathrm L}}(\mathrm{Tok}(c))\in\mathbb R^{m\times4096}$, where $\mathrm{Embed}_{\theta_{\mathrm L}}$ denotes the tokenwise embedding operation. We then combine the visual and language token embeddings by inserting the visual embeddings immediately after the beginning-of-sequence token, yielding the multimodal embedding sequence $H_t=\operatorname{Concat}_{\mathrm{rows}}(H_t^{\mathrm{lang}}[1,:],H_t^{\mathrm{img}},H_t^{\mathrm{lang}}[2:m,:])\in\mathbb R^{(m+256)\times4096}$, where $\operatorname{Concat}_{\mathrm{rows}}$ concatenates its arguments along the row dimension, $H_t^{\mathrm{lang}}[1,:]$ is treated as a single row, and $H_t^{\mathrm{lang}}[2:m,:]$ denotes the remaining rows. The resulting multimodal embedding sequence $H_t$ is then fed into Llama~2 7B~\citep{touvron2023llama} for autoregressive action-token prediction. Specifically, let $\mathrm{Llama}_{\theta_{\mathrm L}}:\mathbb R^{(m+256)\times4096}\times\mathcal V^\ast\to\mathcal P(\mathcal V)$ denote its next-token predictor, including the embedding of preceding action tokens, where $\mathcal P(\mathcal V)$ denotes the set of probability distributions over $\mathcal V$. For an action-token sequence $s=(s_1,\ldots,s_7)\in\mathcal V^7$, the model defines $p_\theta(s\mid o_t,c)=\prod_{j=1}^{7}\mathrm{Llama}_{\theta_{\mathrm L}}(H_t,s_{<j})(s_j)$, where $s_{<j}=(s_1,\ldots,s_{j-1})$ denotes the preceding action tokens and is empty for $j=1$. The resulting action-token sequence $s_t=(s_{t,1},\ldots,s_{t,7})\in\mathcal V^7$ is converted into a continuous action vector $a_t=\mathrm{Detokenizer}(s_t)\in\mathcal A$, where $\mathrm{Detokenizer}:\mathcal V^7\to\mathcal A$ denotes OpenVLA's fixed action-token decoding and dataset-specific unnormalization procedure. Thus, the model output is $y_t=(s_t,a_t)\in\mathcal Y$. The resulting seven-dimensional vector $a_t$ is the robot action command, which is then executed through the robot's control interface.
\paragraph{Pre-training.}
We initialize $\theta$ from the pretrained Prismatic VLM~\citep{karamcheti2024prismatic}, which integrates DINOv2, SigLIP, and Llama~2 7B and is trained on vision-language data. We do not elaborate on this upstream VLM training. Instead, OpenVLA pre-training further optimizes $\theta$ using robot demonstration data from Open X-Embodiment~\citep{kim2025openvla}. Following the unified formulation, let $\mathcal D_{\mathrm{pre}}=\{z_i^{\mathrm{pre}}\}_{i=1}^{N_{\mathrm{pre}}}$, where $z_i^{\mathrm{pre}}=(o_i,c_i,a_i)$ consists of an observation $o_i\in\mathcal O$, a language instruction $c_i\in\mathcal C$, and a ground-truth robot action $a_i\in\mathcal A$, and $N_{\mathrm{pre}}$ is the number of training examples. Each action $a_i$ is normalized and discretized using OpenVLA's fixed action tokenizer to obtain the ground-truth action-token sequence $s_i=(s_{i,1},\ldots,s_{i,7})\in\mathcal V^7$. Given the multimodal embedding sequence $H_i$ constructed from $(o_i,c_i)$ as described above, the per-example pre-training cross-entropy loss is defined as
\begin{equation}
\ell_{\mathrm{pre}}(\theta,z_i^{\mathrm{pre}})=-\frac{1}{7}\sum_{j=1}^{7}\log\mathrm{Llama}_{\theta_{\mathrm L}}(H_i,s_{i,<j})(s_{i,j}),
\label{eq:ovpre}
\end{equation}
where $s_{i,<j}=(s_{i,1},\ldots,s_{i,j-1})$ denotes the ground-truth preceding action tokens, with $s_{i,<1}$ being empty. The loss uses teacher forcing to supervise the seven action tokens. Following Eq.~\eqref{eq:pretr}, we minimize the empirical objective $\widehat{\mathcal L}_{\mathrm{pre}}(\theta)=\frac{1}{N_{\mathrm{pre}}}\sum_{i=1}^{N_{\mathrm{pre}}}\ell_{\mathrm{pre}}(\theta,z_i^{\mathrm{pre}})$ and denote the resulting pretrained VLA parameters by $\theta_{\mathrm{pre}}$.
\paragraph{Post-training.} Starting from the pretrained VLA parameters $\theta^{(0)}=\theta_{\mathrm{pre}}$, OpenVLA performs supervised fine-tuning (SFT) on downstream robot demonstrations $\mathcal D_{\mathrm{off}}=\{z_i^{\mathrm{off}}\}_{i=1}^{N_{\mathrm{off}}}$ (e.g., task-specific demonstrations for placing a carrot in a bowl, moving an instructed object onto a plate, and wiping a table, collected using the Franka-Tabletop and Franka-DROID robot setups)~\citep{kim2025openvla}, where $z_i^{\mathrm{off}}=(o_i,c_i,a_i)$ follows the same data format as in pre-training. SFT uses the same teacher-forced action-token cross-entropy loss as pre-training, but optimizes it on downstream demonstrations. Since only offline data are used, we set $\alpha_k=0$ for all $k$, reducing Eq.~\eqref{eq:postr} to $\widehat{\mathcal L}_{\mathrm{post}}^{(k)}(\theta)=\frac{1}{N_{\mathrm{off}}}\sum_{i=1}^{N_{\mathrm{off}}}\ell_{\mathrm{off}}^{(k)}(\theta,z_i^{\mathrm{off}})$, where $\ell_{\mathrm{off}}^{(k)}$ has the same functional form as $\ell_{\mathrm{pre}}$. Optimizing this objective yields the post-trained parameters $\theta_{\mathrm{post}}=\theta^{(K)}$.
\paragraph{Deployment.} Following the deployment procedure in the unified algorithm, OpenVLA uses greedy decoding to generate action tokens. Specifically, at each step $t$, given the multimodal embedding sequence $H_t$ constructed from $(o_t,c)$, the $j$-th action token is selected as
\begin{equation}
s_{t,j}=\arg\max_{v\in\mathcal V}\mathrm{Llama}_{\theta_{\mathrm L}^{\mathrm{post}}}(H_t,s_{t,<j})(v),\quad j=1,\ldots,7,
\end{equation}
where $\theta_{\mathrm L}^{\mathrm{post}}$ denotes the language-model parameters within $\theta_{\mathrm{post}}$, and $s_{t,<j}=(s_{t,1},\ldots,s_{t,j-1})$ denotes the previously generated action tokens, with $s_{t,<1}$ being empty. The resulting sequence $s_t$ is detokenized into $a_t=\mathrm{Detokenizer}(s_t)$, yielding $y_t=(s_t,a_t)$. Since greedy decoding and action detokenization are deterministic, $f_{\theta_{\mathrm{post}}}(\cdot\mid o_t,c)$ assigns probability one to $y_t$, and the action-realization kernel $q_{\theta_{\mathrm{post}}}$ deterministically extracts $a_t$ from $y_t$.
\section{World Action Models (WAMs)}
\label{sec:wamod}

While conventional vision-language-action (VLA) models primarily learn to predict robot actions from visual and language inputs, world-action models (WAMs) additionally model future environment evolution by jointly predicting future observations and robot actions. We discuss DreamZero~\citep{ye2026world} in detail below.

\paragraph{Model Architecture.}
We specialize the unified formulation to DreamZero~\citep{ye2026world}, representing the model as $f_\theta:\mathcal T\times\mathcal C\to\mathcal P(\mathcal Y)$, where $\mathcal T$ is the interaction history space, $\mathcal C$ is the language instruction space, and $\mathcal Y=\mathcal I^F\times\mathcal A^H$ is the joint video-action output space. Here, $\mathcal I\subseteq\mathbb R^{h\times w\times3}$ denotes the RGB image space, $\mathcal A\subseteq\mathbb R^d$ denotes the robot action space, and $F,H\ge1$ are the numbers of predicted video frames and robot actions, respectively. At step $t$, the observation $o_t=(I_t,p_t)\in\mathcal O=\mathcal I\times\mathcal S$ consists of an image $I_t\in\mathcal I$ and proprioceptive state $p_t\in\mathcal S\subseteq\mathbb R^{d_{\mathrm s}}$. DreamZero conditions on the available visual history $I_{\le t}=(I_0,\ldots,I_t)$, current proprioceptive state $p_t$, and language instruction $c\in\mathcal C$. The model builds upon Wan2.1-I2V-14B~\citep{wan2025wan} and employs a joint video-action Diffusion Transformer (DiT)~\citep{peebles2023scalable}. Let $\theta=(\theta_{\mathrm D},\theta_{\mathrm S},\theta_{\mathrm E},\theta_{\mathrm A})$ denote the parameters of the DiT backbone, state encoder, action encoder, and action decoder, respectively.

\begin{itemize}

\item \textbf{Encoding.}
First, DreamZero encodes the visual history, language instruction, and proprioceptive state into a conditioning representation. Let $I_t^{\mathrm{ctx}}$ denote the finite visual context retained from $I_{\le t}$. The fixed video VAE encoder $\mathrm{VAEEnc}$ maps this context into video latents, while the fixed image encoder $\mathrm{ImageEnc}:\mathcal I\to\mathbb R^{257\times1280}$ extracts features from the current image $I_t$. The fixed text encoder $\mathrm{TextEnc}:\mathcal C\to\mathbb R^{m\times4096}$ processes the language instruction, where $m\le512$ denotes the number of text tokens. The state encoder $\mathrm{StateEnc}_{\theta_{\mathrm S}}:\mathcal S\to\mathbb R^{5120}$ embeds the proprioceptive state, including padding to 64 dimensions. These representations form the conditioning tuple $C_t=(\mathrm{VAEEnc}(I_t^{\mathrm{ctx}}),\mathrm{ImageEnc}(I_t),\mathrm{TextEnc}(c),\mathrm{StateEnc}_{\theta_{\mathrm S}}(p_t))$. Separately, the action encoder $\mathrm{ActionEnc}_{\theta_{\mathrm E}}:\mathbb R^{H\times32}\times[0,1]\to\mathbb R^{H\times5120}$ embeds the action variables and flow time for joint processing within the DiT.

\item \textbf{Joint Video-Action DiT.}
Second, DreamZero jointly models future video and robot actions in a latent space. Let $\mathcal Z=\mathbb R^{2\times16\times h_{\mathrm v}\times w_{\mathrm v}}$ denote the video latent chunk space, where $h_{\mathrm v}=h/8$ and $w_{\mathrm v}=w/8$, and define the joint latent space $\mathcal X=\mathcal Z\times\mathbb R^{H\times32}$. Given future video frames $\mathbf I\in\mathcal I^F$ and corresponding robot actions $\mathbf a\in\mathcal A^H$ spanning the same time interval, the clean joint sample is $x_1=(\mathrm{VAEEnc}(\mathbf I),\mathrm{Normalize}(\mathbf a))\in\mathcal X$, where $\mathrm{Normalize}:\mathcal A^H\to\mathbb R^{H\times32}$ performs action normalization and padding. The video VAE includes the temporal processing required to represent each predicted chunk using two latent frames. With Gaussian noise $x_0\sim\mathcal N(0,I_{\dim\mathcal X})$ and flow time $s\in[0,1]$, the intermediate joint state is $x_s=(1-s)x_0+sx_1$, where the joint space is identified with its vectorized Euclidean representation. The DiT predicts the corresponding velocity through
\begin{equation}
v_\theta(x_s,s,C_t)=\mathrm{DiT}_\theta(x_s,s,C_t)\in\mathcal X,
\label{eq:dzvel}
\end{equation}
where $\mathrm{DiT}_\theta$ includes the action encoder, shared transformer backbone, and modality-specific input and output transformations.

\item \textbf{Generation and Decoding.}
Third, DreamZero generates future video frames and robot actions through the learned joint velocity field. Starting from Gaussian noise, the conditional flow produces a generated joint latent $\widehat{x}_1=(\widehat{z}_1,\widehat{b}_1)\in\mathcal X$. The video component is decoded through the fixed video VAE decoder $\mathrm{VAEDec}$, while the action component is converted into executable robot actions through the fixed unnormalization operation $\mathrm{Unnormalize}:\mathbb R^{H\times32}\to\mathcal A^H$. The resulting output is $y=(\mathbf I',\mathbf a')\in\mathcal Y$, where $\mathbf I'=\mathrm{VAEDec}(\widehat{z}_1)\in\mathcal I^F$ and $\mathbf a'=\mathrm{Unnormalize}(\widehat{b}_1)\in\mathcal A^H$. Thus, DreamZero jointly generates predicted future observations and their corresponding robot actions.

\end{itemize}

\paragraph{Pre-training.}
DreamZero~\citep{ye2026world} is pre-trained on robot interaction trajectories from approximately 500 hours of AgiBot G1 teleoperation data and the DROID dataset for Franka, with separate models trained for the two embodiments. Following the unified formulation, let $\mathcal D_{\mathrm{pre}}=\{z_i^{\mathrm{pre}}\}_{i=1}^{N_{\mathrm{pre}}}$, where $z_i^{\mathrm{pre}}=(\tau^{(i)},c^{(i)})$ consists of an interaction trajectory $\tau^{(i)}=(o_0^{(i)},a_0^{(i)},\ldots,a_{T_i-1}^{(i)},o_{T_i}^{(i)})$ and its corresponding language instruction $c^{(i)}$. Each trajectory is divided into $M$ temporally aligned video-action chunks. For chunk $j$, the clean joint sample is $x_1^{(j)}=(\mathrm{VAEEnc}(\mathbf I^{(j)}),\mathrm{Normalize}(\mathbf a^{(j)}))\in\mathcal X$. Independently sampling Gaussian noise $x_0^{(j)}\sim\mathcal N(0,I_{\dim\mathcal X})$ and flow time $s_j\sim\mathcal U(0,1)$, we construct $x_{s_j}^{(j)}=(1-s_j)x_0^{(j)}+s_jx_1^{(j)}$, with target velocity $v^{(j)}=x_1^{(j)}-x_0^{(j)}$. DreamZero uses the same flow time for the video and action components within each chunk. During training, teacher forcing conditions each noisy chunk on the preceding clean chunks. Let $\widetilde C_{i,j}$ denote this teacher-forced conditioning representation, incorporating the preceding clean context, language instruction, and current proprioceptive state from $z_i^{\mathrm{pre}}$. The per-example joint flow-matching loss is defined as
\begin{equation}
\ell_{\mathrm{pre}}(\theta,z_i^{\mathrm{pre}})=\mathbb E_{\{s_j,x_0^{(j)}\}_{j=1}^{M}}\left[\frac{1}{M}\sum_{j=1}^{M}w(s_j)\left\|v_\theta(x_{s_j}^{(j)},s_j,\widetilde C_{i,j})-v^{(j)}\right\|_2^2\right],
\label{eq:dzpre}
\end{equation}
where $w(s_j)>0$ is a predefined flow-time weighting function. The loss jointly supervises video and action velocity prediction, allowing the shared DiT to learn environment dynamics and robot control. Following Eq.~\eqref{eq:pretr}, we minimize the empirical objective $\widehat{\mathcal L}_{\mathrm{pre}}(\theta)=\frac{1}{N_{\mathrm{pre}}}\sum_{i=1}^{N_{\mathrm{pre}}}\ell_{\mathrm{pre}}(\theta,z_i^{\mathrm{pre}})$ and denote the resulting pretrained parameters by $\theta_{\mathrm{pre}}$. During optimization, the DiT backbone, state encoder, action encoder, and action decoder are updated, while the video VAE, image encoder, and text encoder remain frozen.

\paragraph{Post-training.}
DreamZero~\citep{ye2026world} adapts the pretrained WAM to downstream manipulation tasks through offline supervised fine-tuning (SFT). Starting from $\theta^{(0)}=\theta_{\mathrm{pre}}$, the model is fine-tuned separately on task-specific robot demonstrations, including shirt folding, fruit packing, and table bussing. For each task, let $\mathcal D_{\mathrm{off}}=\{z_i^{\mathrm{off}}\}_{i=1}^{N_{\mathrm{off}}}$, where $z_i^{\mathrm{off}}=(\tau_i,c_i)$ follows the same data format as pre-training. SFT retains the joint flow-matching objective in Eq.~\eqref{eq:dzpre}, using the same noise interpolation, velocity supervision, and teacher forcing. Since only offline data are used, we set $\alpha_k=0$ in Eq.~\eqref{eq:postr}, yielding $\widehat{\mathcal L}_{\mathrm{post}}^{(k)}(\theta)=\frac{1}{N_{\mathrm{off}}}\sum_{i=1}^{N_{\mathrm{off}}}\ell_{\mathrm{off}}^{(k)}(\theta,z_i^{\mathrm{off}})$, where $\ell_{\mathrm{off}}^{(k)}$ has the same functional form as $\ell_{\mathrm{pre}}$. The DiT backbone, state encoder, action encoder, and action decoder are fine-tuned, while the video VAE, image encoder, and text encoder remain frozen. The model is optimized for 50,000 steps on each downstream task, producing task-specific post-trained parameters $\theta_{\mathrm{post}}=\theta^{(K)}$.

\paragraph{Deployment.}
DreamZero~\citep{ye2026world} uses the post-trained parameters $\theta_{\mathrm{post}}$ to jointly generate future video frames and robot actions for closed-loop control. Given the available interaction history $\tau_{\le t}$ and language instruction $c$, the model constructs the conditioning representation $C_t$ from the observed visual history, current proprioceptive state, and language instruction. To generate a new video-action chunk, DreamZero samples $x_0\sim\mathcal N(0,I_{\dim\mathcal X})$ and numerically integrates the learned velocity field according to
\begin{equation}
\frac{\mathrm d x_s}{\mathrm ds}=v_{\theta_{\mathrm{post}}}(x_s,s,C_t),\qquad x_s\big|_{s=0}=x_0,\qquad s\in[0,1],
\label{eq:dzode}
\end{equation}
obtaining the generated joint latent $\widehat x_1=(\widehat z_1,\widehat b_1)$. The video and action components are decoded as $\mathbf I'=\mathrm{VAEDec}(\widehat z_1)$ and $\mathbf a'=\mathrm{Unnormalize}(\widehat b_1)$, yielding a new joint output $y=(\mathbf I',\mathbf a')\in\mathcal Y$. For real-time control, DreamZero executes action chunks asynchronously while generating subsequent chunks using updated observations. At control step $t$, let $y_t$ denote the available joint chunk selected for execution, which may have been generated at an earlier step. The deployed kernel $f_{\theta_{\mathrm{post}}}(\cdot\mid\tau_{\le t},c)$ represents the conditional distribution of this available output, including asynchronous generation and scheduling. The action-realization kernel $q_{\theta_{\mathrm{post}}}(\cdot\mid y_t,\tau_{\le t},c)$ represents the selection of the executable action from the available chunk, including action smoothing when applied. Together, $f_{\theta_{\mathrm{post}}}$ and $q_{\theta_{\mathrm{post}}}$ induce the deployed policy $\pi_{\theta_{\mathrm{post}}}$ through Eq.~\eqref{eq:polic}. After executing $a_t$, the robot receives a new observation $o_{t+1}=(I_{t+1},p_{t+1})$ and updates the interaction history as $\tau_{\le t+1}=(\tau_{\le t},a_t,o_{t+1})$. The updated observations condition subsequent video-action generation, enabling closed-loop interaction.

\section{Discussion}
We present the first architecture-agnostic probabilistic formulation based on stochastic kernels that explicitly connects pre-training, post-training, and closed-loop robot interaction for embodied foundation models. By relating heterogeneous model outputs to task-conditioned robot policies, this formulation provides a common probabilistic language for understanding embodied models with different architectures, output representations, and learning objectives. More importantly, it clarifies how pre-training and post-training shape the policies deployed in closed-loop interaction. Future work may build upon this framework to investigate how different pre-training and post-training strategies affect policy-induced distribution shifts and generalization across tasks, environments, and embodiments.

\newpage
\bibliographystyle{plainnat}
\bibliography{references}

\end{document}